\pdfoutput=1
\documentclass[11pt,a4paper]{article}
\usepackage[hyperref]{emnlp2021}

\usepackage[T1]{fontenc}
\usepackage[utf8]{inputenc}

\usepackage{enumerate}
\usepackage{enumitem}
\usepackage{times}
\usepackage{latexsym}

\usepackage{microtype}
\usepackage{graphicx, textcomp, comment, hyperref, tabularx, natbib, amsmath, enumitem, caption, amsfonts, xspace}
\usepackage{booktabs,tabularx,multirow}
\newcolumntype{Y}{>{\centering\arraybackslash}X}

\usepackage{tikz}
\newcommand*\circled[2]{\tikz[baseline=(char.base)]{
    \node[shape=circle, draw, inner sep=1pt, scale=0.825, fill=#2] (char) {#1};}}
\newcommand*\dyadicmodel[1]{\circled{#1}{blue!20}}
\newcommand*\systemicmodel[1]{\circled{#1}{green!20}}
\newcommand*\ceilingmodel[0]{\circled{C}{orange!20}}

\usepackage{inconsolata} % for texttt{}
\usepackage{cleveref}

\crefname{section}{\S}{\S\S}
\Crefname{section}{\S}{\S\S}
\crefname{table}{Tab.}{}
\crefname{figure}{Fig.}{}
\crefname{algorithm}{Algorithm}{}
\crefname{equation}{eq.}{}
\crefname{appendix}{App.}{}
\crefname{thm}{Theorem}{}
\crefname{prop}{Proposition}{}
\crefname{cor}{Corollary}{}
\crefname{observation}{Observation}{}
\crefname{assumption}{Assumption}{}
\crefformat{section}{\S#2#1#3}

\usepackage{bbm}
\usepackage{todonotes}
\makeatletter
\newcommand*\iftodonotes{\if@todonotes@disabled\expandafter\@secondoftwo\else\expandafter\@firstoftwo\fi}  % defines \iftodonotes{<true>}{<false>}, thanks to https://tex.stackexchange.com/questions/126559/conditional-based-on-packageoption
\makeatother

\newcommand{\SetSize}[1]{\lvert #1\rvert}

\def\calE{{\mathcal{E}}}

\def\calN{{\mathcal{N}}}

\newcommand{\crl}[1]{\left\{#1\right\}}

\newcommand{\defn}[1]{\textbf{#1}}

\newcommand{\Gdi}{G_{\setminus d_i}}
\newcommand{\Ndi}{\mathcal{N}_{\setminus d_i}}
\newcommand{\Edi}{\mathcal{E}_{\setminus d_i}}

\newcommand{\Edilab}{\Edi^\text{lab}}

\newcommand{\ConflictWiki}{ConflictWiki\xspace}
\newcommand{\Entity}[1]{\textit{#1}}
\newcommand{\Conflict}[1]{\textit{#1}}
\newcommand{\WikiSection}[1]{\textit{#1}}
\newcommand{\TechnicalTag}[1]{\texttt{#1}\xspace}

\newcommand{\FOne}{\text{F1}\xspace}

\newcommand{\ally}{\textsc{ally}\xspace}
\newcommand{\enemy}{\textsc{enemy}\xspace}
\newcommand{\allies}{\textsc{allies}\xspace}
\newcommand{\enemies}{\textsc{enemies}\xspace}

\definecolor{good}{RGB}{67, 148, 61}
\definecolor{bad}{RGB}{173, 26, 26}

\usepackage[group-separator={,}]{siunitx}

\usepackage{tipa}

\newcommand{\ucambridge}{\normalfont \text{\textipa{D}}}
\newcommand{\ethz}{\text{\normalfont \textipa{Q}}}
\newcommand{\mitinst}{\normalfont \text{\textipa{@}}}
\newcommand{\epfl}{\normalfont \text{\textipa{N}}}

\title{Classifying Dyads for Militarized Conflict Analysis}

\author{
Niklas Stoehr$^{\ethz}$~\;~Lucas Torroba Hennigen$^{\mitinst}$~\;~\;Samin Ahbab \\ ~\;~\textbf{Robert West}$^{\epfl}$~\;~\textbf{Ryan Cotterell}$^{\ethz, \ucambridge}$
\\
$^{\ethz}$ETH Z{\"u}rich \quad $^{\mitinst}$MIT \quad $^{\epfl}$EPFL \quad $^{\ucambridge}$University of Cambridge\\
\href{mailto:niklas.stoehr@inf.ethz.ch}{\texttt{niklas.stoehr@inf.ethz.ch}}~\;~ \href{mailto:lucastor@mit.edu}{\texttt{lucastor@mit.edu}} 
~\;~\href{mailto:saminahbab0@gmail.com}{\texttt{saminahbab0@gmail.com}}\\
\href{mailto:robert.west@epfl.ch}{\texttt{robert.west@epfl.ch}}~\;~ \href{mailto:ryan.cotterell@inf.ethz.ch}{\texttt{ryan.cotterell@inf.ethz.ch}}
}

\date{}

\begin{document}

\maketitle
\begin{abstract}

Understanding the origins of militarized conflict is a complex, yet important undertaking. Existing research seeks to build this understanding by considering bi-lateral relationships between entity pairs (dyadic causes) and multi-lateral relationships among multiple entities (systemic causes).  The aim of this work is to compare these two causes in terms of how they correlate with conflict between two entities. We do this by devising a set of textual and graph-based features which represent each of the causes. The features are extracted from Wikipedia and modeled as a large graph. Nodes in this graph represent entities connected by labeled edges representing ally or enemy-relationships. This allows casting the problem as an edge classification task, which we term dyad classification. We propose and evaluate classifiers to determine if a particular pair of entities are allies or enemies. Our results suggest that our systemic features might be slightly better correlates of conflict. Further, we find that Wikipedia articles of allies are semantically more similar than enemies.\footnote{Our dataset can be explored in an interactive dashboard: \href{https://conflict-ai.github.io/conflictwiki}{https://conflict-ai.github.io/conflictwiki}.
Data, code and documentations are provided at \href{https://github.com/conflict-ai/conflictwiki}{https://github.com/conflict-ai/conflictwiki}.}\looseness=-1
\end{abstract}

\section{Introduction}
\label{sec:introduction}

Researchers have long sought to understand the underlying causes of militarized conflict.
The origins of conflict can be broadly categorized as either \defn{dyadic} or \defn{systemic}. Dyadic pertains to entity-specific idiosyncrasies, competing ideologies \cite{leader_maynard_ideology_2019}, e.g., dissimilar political systems \cite{rousseau_assessing_1996}, and power differentials, e.g., economic and demographic capabilities \cite{geller_power_1993}. 
The \href{https://en.wikipedia.org/wiki/Mali_War}{\Conflict{Mali War}} is an example of a conflict to which a dyadic cause has been attributed: It was spawned from differing cultural and ideological identities between the \href{https://en.wikipedia.org/wiki/National_Movement_for_the_Liberation_of_Azawad}{\Entity{Azawad Liberation Movement}} and the \href{https://en.wikipedia.org/wiki/Mali}{\Entity{Malian government}} \cite{chauzal_roots_2015}. Throughout this paper, we use the term dyad to not only denote conflictual entity pairs (\enemies) but also cooperative pairs (\allies) in a conflict \cite{geller_power_1993}.\looseness=-1

\begin{figure}[t]
    \centering
    \includegraphics[width=1.0\linewidth]{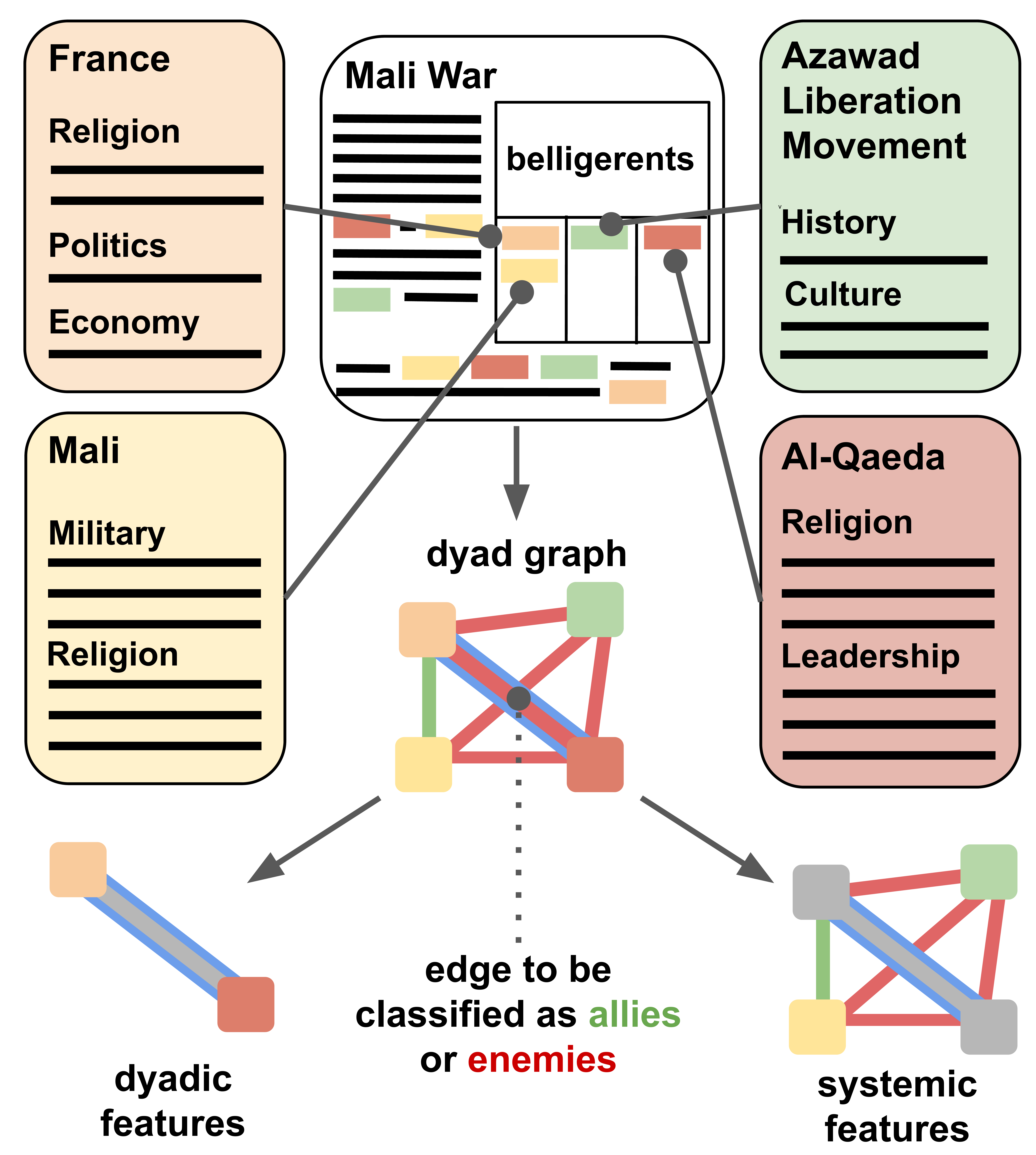} 
    \caption{Overview of our approach. From \href{https://en.wikipedia.org/wiki/Category:Conflicts}{Wikipedia articles on conflicts}, we extract the \href{https://en.wikipedia.org/wiki/Template:Infobox_military_conflict}{belligerents table in the infobox}. This allows constructing a dyad graph in which nodes represent entities connected by labeled edges representing \ally or \enemy-relationships. We then compare dyadic and systemic features in terms of how effective they are at classifying two entities as allies or enemies. We term this task dyad classification.}
    \label{fig:overview}
\end{figure}

\begin{figure*}[t]
    \centering
    \includegraphics[width=1.0\textwidth]{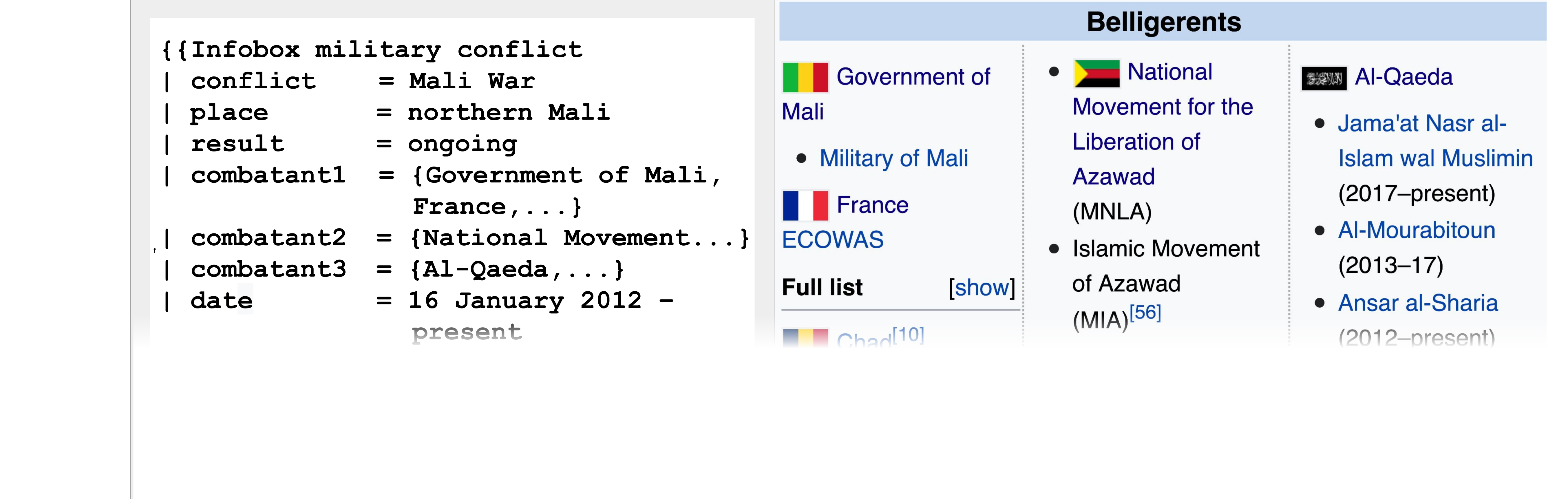} 
    \caption{Example of the template used for displaying belligerents in the infobox of Wikipedia conflict articles. The left hand side shows what the Wikipedia \href{https://en.wikipedia.org/wiki/Template:Infobox_military_conflict}{infobox template} and the included metadata look like. Note that there may be two or more \TechnicalTag{combatant} tags indicating opposing conflict parties. The right hand side of the figure shows the \href{https://en.wikipedia.org/wiki/Mali_War}{relevant section of the infobox} that contains the belligerents.
}\label{fig:infobox}
\end{figure*}

A systemic cause, on the other hand, is a cause based on the broader relationship network involving a larger set of entities \cite{sweeney_dyadic_2004, rasler_systemic_2010}.
For instance, the intervention of \href{https://en.wikipedia.org/wiki/France}{\Entity{France}} in the \href{https://en.wikipedia.org/wiki/Mali_War}{\Conflict{Mali War}} may be said to have had a systemic cause as its origins may be partly attributed to NATO's close diplomatic ties with the Economic Community of West African States \cite[ECOWAS;][]{francis_regional_2013}. Determining a systemic cause may be aided by the analysis of a graph that encodes the relationships between the entities in the conflict. Another example of a systemic origin of conflict is the ancient proverb ``the enemy of my enemy is my friend'', which is also known as the \defn{structural balance theory} \cite{heider_psychology_1946, cartwright_structural_1956}.\looseness=-1

In this paper, we construct textual and graph-based features that encode dyadic and systemic correlates of conflict. We take this approach since establishing causality from our data is a highly complex endeavor, so we focus on correlates instead. Our approach uses Wikipedia data to compare the ability of classifiers trained using these features to predict whether two entities are allies or enemies. This is illustrated at a high level in \cref{fig:overview}. We then perform an ablation study by systematically leaving out dyadic and systemic features to ascertain to what degree these features correlate with whether two entities are enemies or allies in a conflict. Our systemic model obtains an $\FOne$ score of $0.917$ and our dyadic model obtains an $\FOne$ score of $0.873$. If one believes our features to be representative dyadic and systemic correlates, then this result provides support for the claim that, in aggregate, systemic causes may play a slightly larger role. Moreover, we also find that articles of allies are semantically more similar than enemies.

\section{Dyadic and Systemic Features}
\label{sec:operationalization}

The larger scientific mission of this paper is to investigate whether, when analyzing a large corpus of conflicts, dyadic or systemic correlates of conflict are more prominent.
To carry out such a study, we construct features that encode the notions of dyadic and systemic and train classifiers to predict whether two entities are enemies or allies using these features. The classifiers are designed to operate on a \defn{dyad graph}, an undirected graph where each node corresponds to an entity and each edge corresponds to the relationship between two entities. This allows casting the problem as an edge classification task, which we term \defn{dyad classification}. In this section we describe the features accessible by both models, deferring the construction of the dyad graph from Wikipedia to \cref{sec:dataset-building} and the actual technical implementation of the models to \cref{sec:experimental-setup}.

\paragraph{Notation.} Let $G = (\mathcal{N}, \mathcal{E})$ be the dyad graph consisting of entity nodes $\calN$ and labeled relationship edges $\calE$. This graph can be equivalently represented by the set of all dyads $D = \crl{ d_i }_{i = 1}^{\SetSize{D}}$. Each dyad $d_i = (u_i, v_i, e_i, y_i)$ consists of two entities $u_i, v_i \in \mathcal{N}$ connected by an edge $e_i = (u_i, v_i) \in \mathcal{E}$, which is labeled as $y_i \in \crl{\allies, \enemies}$.

\begin{figure*}[t]
    \centering
    \includegraphics[width=1.0\textwidth]{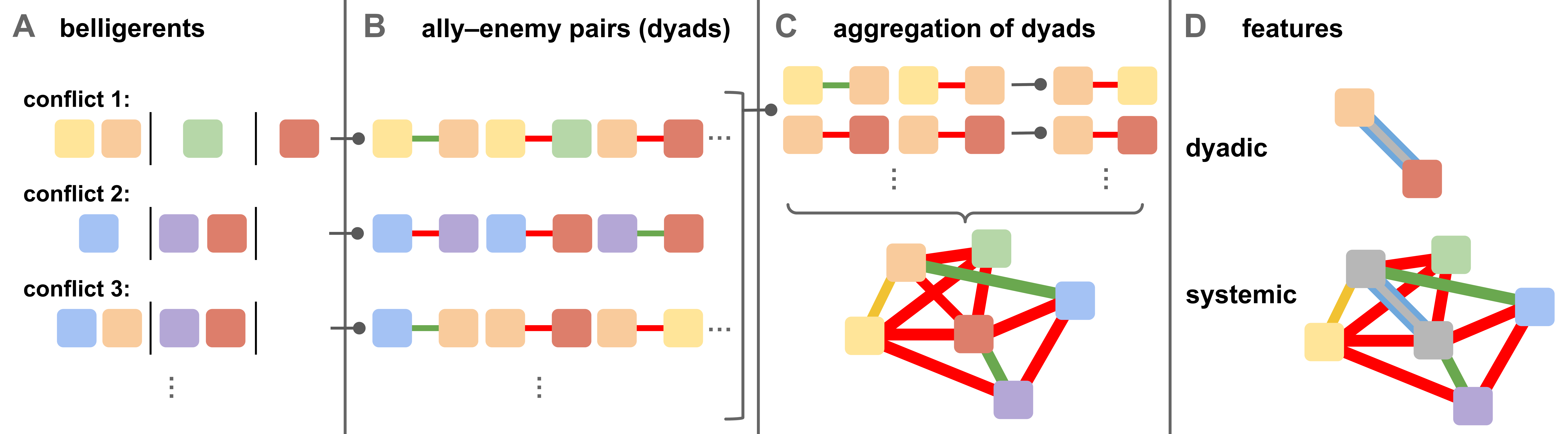}
    \caption{Construction of dyad graph; (A) Entities in each conflict are partitioned into belligerents; (B) We construct entity pairs (dyads) from all combinations of belligerents in a conflict; (C) We aggregate dyads across conflicts into a graph where nodes are entities and edges are conflicts;
    (D) When considering dyadic features, we only expose to the model the dyad that is meant to be classified, but when considering systemic features, we expose the graph information of everything but the dyad to be classified.
    }
    \label{fig:task_construction}
\end{figure*}

\paragraph{Dyadic features.}

The adjective \defn{dyadic} is derived from the noun dyad, which is the basic unit of a militarized conflict and describes a pair of warring entities \cite{harbom_dyadic_2008}. Throughout this paper, we expand the use of the term dyad to not only denote conflictual entity pairs (\enemies) but also cooperative pairs (\allies) in a given conflict \cite{geller_power_1993}. Dyadic correlates pertain to idiosyncrasies of two entities and their bilateral relationship. This suggests that suitable dyadic features would be any information directly associated with a dyad. Particularly, as dyadic features, we consider the representations of both entities, $u_i$ and $v_i$, and the unlabeled edge between them, $e_i$.\looseness=-1

\paragraph{Systemic features.}
Systemic correlates are contained within the wider network of relationships two entities are embedded in. Hence, we can think of systemic features as those that are exposed by a \defn{restricted dyad graph} $\Gdi$, defined as the dyad graph $G$ minus the dyad $d_i$ that is to be classified:
\begin{align}
    \Gdi &= (\Ndi, \Edi) \\
         &= (\mathcal{N} - \{u_i, v_i\}, \mathcal{E} - \{(u_i, v_i, y_i)\}) \nonumber
\end{align}
Specifically, our systemic features are the representations of neighboring entities $\Ndi$, the representations of their relationships $\Edi$, and the labels of those relationships, which we denote $\Edilab$.

\section{Constructing the Dyad Graph}
\label{sec:dataset-building}

We now turn to the problem of extracting a dyad graph from Wikipedia. We first retrieve conflict, e.g., the \Conflict{Mali War}, and entity, e.g., \Entity{France}, \Entity{Al-Qaeda}, articles from Wikipedia\footnote{We use the English Wikipedia dump released on 25 January 2021.}~(\cref{sec:data_retrieval}) and pre-process the articles to obtain vector representations of articles and their sections (\cref{sec:article_representation}).
The resulting \href{https://conflict-ai.github.io/conflictwiki}{\ConflictWiki} dataset is a collection of articles on militarized conflict and their involved entities. Data, code and documentation are publicly available in an \href{https://conflict-ai.github.io/conflictwiki}{interactive dashboard}. A subset of the dyad graph is shown in \cref{fig:network}.

\subsection{Data Retrieval}
\label{sec:data_retrieval}

To obtain conflict articles, we first extract all articles from the Wikipedia subcategory \href{https://en.wikipedia.org/wiki/Category:21st-century_conflicts_by_year}{Category:21st-century conflicts by year}, offering a collection of all conflict articles from 2001 to 2021. We then recursively extract all articles in all of its subcategories up to a depth of 4 levels. While this procedure ensures wide coverage, it includes various articles which do not describe militarized conflicts but instead conflict entities, political figures, movements or geographic locations. 
For this reason, we filter our selection based on a precise militarized conflict criterion---we discard all articles that do not feature at least two belligerents as indicated by the tags \TechnicalTag{combatant1} and \TechnicalTag{combatant2} in their infobox (see \href{https://en.wikipedia.org/wiki/Template:Infobox_military_conflict}{Template:Infobox military conflict} and \cref{fig:infobox}). 
Due to inconsistencies in the usage of tags, our extraction steps require a considerable number of regular expressions.\footnote{The regular expressions find links to redirect pages and mentions of entities within the infobox.} The whole procedure leaves us with \num{1145} annotated militarized conflicts over a period of 20 years.\looseness=-1

To obtain the Wikipedia articles for all entities involved in a conflict, we consider the \TechnicalTag{combatant} tags in each conflict article's infobox (see \href{https://en.wikipedia.org/wiki/Template:Infobox_military_conflict}{Template:Infobox military conflict} and \cref{fig:infobox}).\footnote{If a hyperlink to an entity article leads to a \href{https://en.wikipedia.org/wiki/Wikipedia:Redirect}{redirect page}, we follow the redirection.} There may be two or more \TechnicalTag{combatant} tags, indicating opposing conflict parties which are \href{https://en.wikipedia.org/wiki/Template:Infobox_military_conflict}{displayed as \TechnicalTag{belligerents}} in Wikipedia. Each belligerent comprises one or more entities (states, militias, etc.) that are united as allies in a particular conflict. Entities assigned to different belligerents are enemies in that conflict. Almost all entities are hyperlinked to their own Wikipedia articles, which we retrieve. All together, we gather \num{1245} articles of entities that are involved in at least one conflict.\footnote{Incidentally, we include additional conflict metadata in the dataset we distribute, even though it is not exploited by any of our models. Specifically, we extract the \TechnicalTag{title} and \TechnicalTag{id} of the conflict, the \TechnicalTag{place} and \TechnicalTag{date} tag for spatio-temporal information as well as the \TechnicalTag{strength}, \TechnicalTag{casualty}, \TechnicalTag{commander} and \TechnicalTag{result} tags. Whenever provided in the entity's infobox, we retrieve auxiliary information on languages, religion, ISO2 code and ideology. We hope that this will be helpful to researchers conducting further work in this area.}

\subsection{Dyad Graph Construction}
\label{sec:task_construction}

The 3-step process for building a dyad graph is depicted in \cref{fig:task_construction}A,B,C. 
The retrieved data yields a set of conflicts. Each one of these conflicts can be thought of as a group of warring factions, which we call \defn{belligerents}. This is illustrated in \cref{fig:task_construction}A. For example, for the \href{https://en.wikipedia.org/wiki/Mali_War}{\Conflict{Mali War}} conflict, we have three sets of belligerents: (1) $\{ \Entity{Mali},\allowbreak \Entity{France} \}$, (2) $\{ \Entity{Azawad Liberation Movement} \}$, and (3) $\{ \allowbreak \Entity{Al-Qaeda} \}$.
Next, we construct a set of ally--enemy pairs for each conflict by taking the Cartesian product of all entities involved in a conflict.
For each pair of entities in a conflict, we take them to be enemies in that conflict if the two entities are in different belligerent sets, and as allies otherwise. This is displayed in \cref{fig:task_construction}B, where green edges represent allies and red edges represent enemies.

\begin{figure}[t]
    \centering
    \includegraphics[width=1.0\columnwidth]{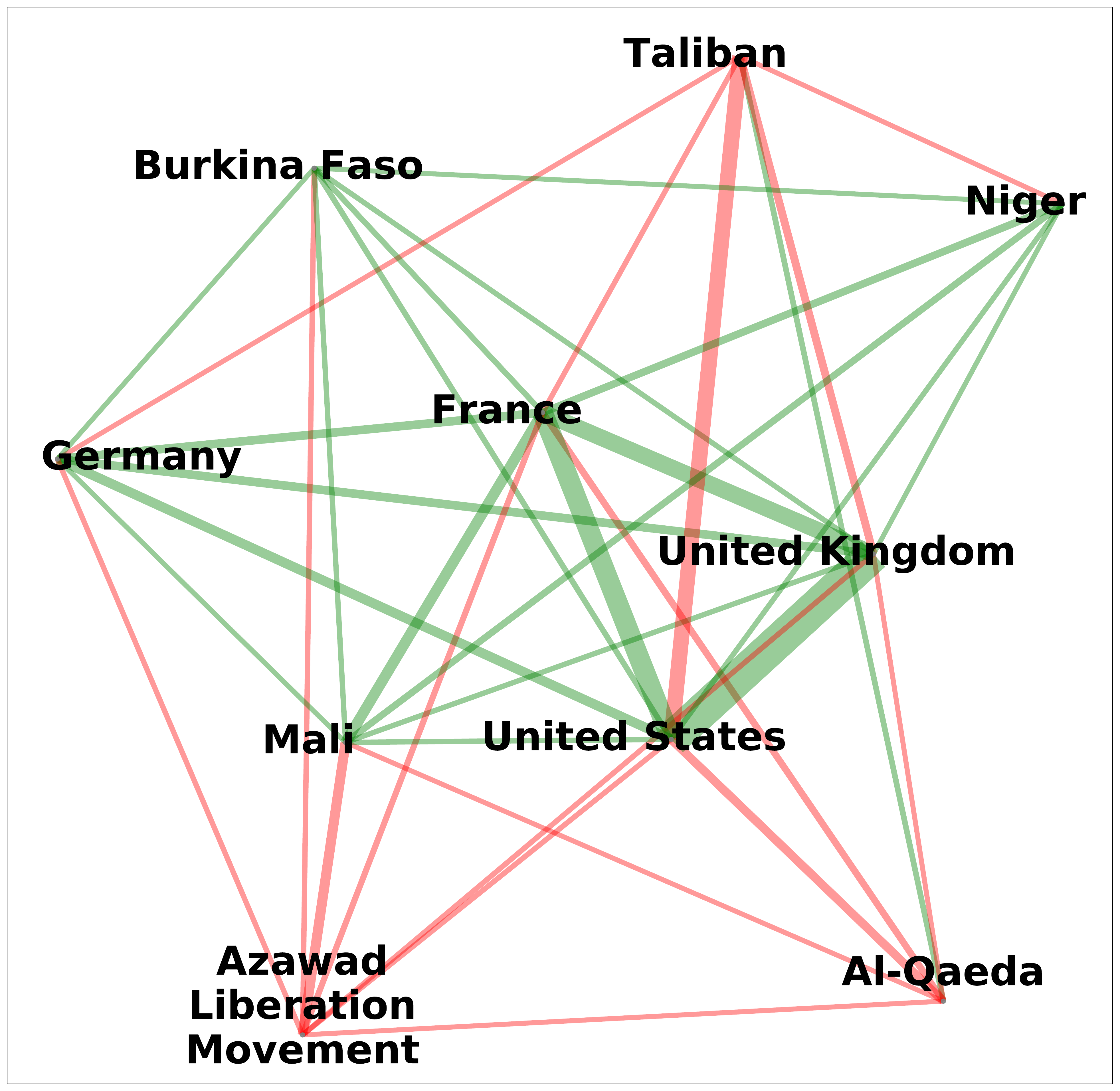}
    \caption{\href{https://conflict-ai.github.io/conflictwiki}{A small subset of the aggregated dyad graph}. Each node represents an entity, each edge represents the relationship (green indicates allies and red indicates enemies) of two entities that participated together in at least one conflict. The edge line width is proportional to the number of conflicts shared by both entities; note that we do this only for illustrative purposes, to emphasize the aggregation of multiple conflict edges into one. The models do not have access to this information.}
    \label{fig:network}
\end{figure}

Next, we aggregate all ally--enemy pairs across conflicts to construct a graph $G = (\calN, \calE)$ as displayed in \cref{fig:task_construction}C.\footnote{Most graph processing is done with the help of the \href{https://networkx.org}{\textit{networkX library}} \cite{networkx}.} The set of nodes $\calN$ represents the set of all entities and the set of edges $\calE$ represents the bilateral relationships between all entities that have been engaged in at least one conflict together, where multiple conflicts between a pair of entities are aggregated into a single edge.
We label an edge as allies if, across all conflicts they partook together, the two entities have been allies strictly more often than enemies; otherwise the edge is labeled as enemies.
Note that entities that do not co-occur in a conflict have no edge between them.
The resulting dyad graph contains a total of \num{26536} ally--enemy edges, with $55\%$ of them being labeled as allies. A subset of it is displayed in \cref{fig:network}.

\section{Experimental Setup}
\label{sec:experimental-setup}

What remains to be discussed is the conversion of articles to a machine-readable representation (\cref{sec:article_representation}), the technical implementation of each of these models, and the baselines and setup of the experiments (\cref{sec:models}).\looseness=-1

\subsection{Article and Section Features}
\label{sec:article_representation}

Having collected the raw data from Wikipedia and constructed the dyad graph, we need to pre-process \ConflictWiki so that it can be used as input to our models.
However, there are two challenges associated with pre-processing the data we retrieved from Wikipedia. The first is that Wikipedia articles for both conflicts and entities can contain explicit mentions of ally--enemy relationships.
For instance, the conflict article \href{https://en.wikipedia.org/wiki/Mali_War}{\Conflict{Mali War}} states\footnote{as of 14 September 2021} that the \href{https://en.wikipedia.org/wiki/National_Movement_for_the_Liberation_of_Azawad}{\Entity{Azawad Liberation Movement}} ``began fighting a campaign against'' the \href{https://en.wikipedia.org/wiki/Mali}{\Entity{Malian government}}, and the entity article     \href{https://en.wikipedia.org/wiki/France}{\Entity{France}} says that the country ``intervened to help the \href{https://en.wikipedia.org/wiki/Malian_Army}{\Entity{Malian Army}}''.
This poses a dilemma: How can we guarantee that a model that takes as input a Wikipedia article is not simply using superficial linguistic cues to regurgitate these relationships? After all, if this is all that a model does, we could hardly attribute its success to whether it is dyadic or systemic! Therefore, for a fair comparison, we need to ensure that our data does not contain explicit mentions of such relationships.\looseness=-1\footnote{We believe that a model that exploits explicit mentions of ally--enemy relationships in the text is better analyzed through the lens of information extraction and machine reading.}

\begin{table}[]
    \centering
    \begin{tabularx}{\linewidth}{@{}lX@{}}
    \toprule 
    Conflict & Top 10 unigrams \\
    \midrule
    \multirow{2}{*}{Mali War} & soldier, town, attack, troop, rebel, group, city, conflict, northern, hostage \\
    \cmidrule(l){2-2}
    \multirow{3}{*}{Mali} & woman, country, coup, population, president, align, region, control, popular, rate \\
    \cmidrule(l){2-2}
    \multirow{3}{*}{France} & world, country, large, region, territory, department, nuclear, language, population, tourist \\
    \cmidrule(l){2-2}
    \multirow{3}{*}{Al-Qaeda} & attack, group, bombing, organization, militant, member, muslim, leader, senior, government \\
    \cmidrule(l){2-2}
    \multirow{3}{*}{Azawad} & movement, city, army, government, control, independence, military, northern, force, fighter \\
    \bottomrule
    \end{tabularx}
    \caption{Top 10 unigrams by tf-idf weighting of different Wikipedia articles; \href{https://en.wikipedia.org/wiki/Mali_War}{\Conflict{Mali War}} conflict article and four involved entities \href{https://en.wikipedia.org/wiki/Mali}{\Entity{Mali}}, \href{https://en.wikipedia.org/wiki/France}{\Entity{France}}, \href{https://en.wikipedia.org/wiki/Al-Qaeda}{\Entity{Al-Qaeda}} and \href{https://en.wikipedia.org/wiki/National_Movement_for_the_Liberation_of_Azawad}{\Entity{Azawad Liberation Movement}}.}
    \label{tab:tfidf_unigrams}
\end{table}

The second challenge is that we must not inadvertently provide more information to our models than the information in our features (\cref{sec:operationalization}). For example, pre-trained representations have been shown to encode a plethora of information that may skew predictions~\citep{kutuzov_tracing_2017, petroni_language_2019, bouraoui_inducing_2020}. Hence, for our experiments we steer away from pre-trained representations and instead learn all parameters of the model from scratch, using only the data they should have access to.

Due to the challenges listed above, we use term frequency inverse-document frequency~\citep[tf-idf;][]{manning-schutze} to compute vector representations of each section\footnote{Since the first section of each article usually has no title, we denote it as \WikiSection{Summary}. Article sections with headers such as \WikiSection{See also}, \WikiSection{Bibliography}, \WikiSection{References}, \WikiSection{Further reading}, \WikiSection{Sources}, \WikiSection{Literature}, \WikiSection{External links}, \WikiSection{Citations}, \WikiSection{Footnotes} and \WikiSection{Notes} are removed.} of every entity and conflict article. We construct two separate corpora for unigram tokens appearing in conflict and entity articles. Next, we pre-process the corpora following several steps: we filter all tokens that are neither nouns nor adjectives and lemmatise all tokens. The last pre-processing steps pertain to the removal of all named entities using \href{https://spacy.io/usage/linguistic-features#named-entities}{spaCy}~\cite{matthew_spacy_2020}. Particularly, we remove context-indicative tokens such as locations (e.g., Mali), dates (e.g., 2012), nationalities (e.g., French), political groups (e.g., Democrats) and organisations (e.g., Al-Qaeda), but keep world religions. Finally, we transform the unigram distribution of each article and each article section into a \num{500}-dimensional tf-idf feature vector. To this end, we filter tokens appearing in more than \num{40}\% and less than \num{1}\% of articles. Then, we select the top \num{500} tokens based on absolute term frequency across the corpus. \cref{tab:tfidf_unigrams} shows the top \num{10} unigrams by tf-idf weighting of conflict and entity articles associated with the \href{https://en.wikipedia.org/wiki/Mali_War}{\Entity{Mali War}}. \looseness=-1

\begin{table*}[t]
    \centering
\begin{tabular}{ccccccccc}
    \toprule
     \cmidrule(lr){3-8}
     & & \multicolumn{3}{c}{Dyadic features} & \multicolumn{3}{c}{Systemic features} &  \\
     \cmidrule(lr){3-5}
     \cmidrule(lr){6-8}
     & & $u_i$ & $v_i$ & $e_i$ & $\Ndi$ & $\Edi$ & $\Edilab$ & F1 score ($\mu {\pm \text{s.d.}}$) \\
     \midrule
     \multirow{4}{*}{Main models} & \dyadicmodel{D} & \checkmark & \checkmark & \checkmark & & & & 0.873 $\pm$ 0.009 \\
     & \systemicmodel{S} & & & & \checkmark & \checkmark & \checkmark &
     0.917 $\pm$ 0.006 \\
     & \ceilingmodel{} & \checkmark & \checkmark & \checkmark & \checkmark & \checkmark & \checkmark & 0.926 $\pm$ 0.008 \\
      & MAJ & & & & & & & 0.649 $\pm$ 0.009 \\
     \midrule
     \multirow{4}{*}{Ablation study} & \dyadicmodel{1} & \checkmark & \checkmark &  & & & & 0.836 $\pm$ 0.013 \\
     & \systemicmodel{2} & & & & \checkmark & & & 0.828 $\pm$ 0.007 \\
     & \systemicmodel{3} & & & & & \checkmark & & 0.779 $\pm$ 0.009 \\
     & \systemicmodel{4} & & & & & & \checkmark & 0.871 $\pm$ 0.005 \\
     \bottomrule
\end{tabular}
    \caption{Mean results and standard deviation of dyad classification task over 10 runs.
    The top half of the table shows the results of our main comparison; the bottom half shows the results of our feature ablation study (\cref{sec:ablation_features}).
    We find that the systemic model ($\FOne = 0.917$) outperforms the dyadic model ($\FOne = 0.873$).
    }
    \label{tab:results_trans_tfidf}
\end{table*}

\subsection{Model Implementation}
\label{sec:models}

We implement the two main models---one that exploits the dyadic features and one with systemic features---alongside a combined model that has access to both features. Recall that, in the dyad graph, every node represents an entity, and edges between entities represent their enemy or ally relationship across one or more conflicts (\cref{sec:task_construction}). We use the tf-idf vectors of entity articles as \defn{node embeddings}, and the average tf-idf vectors across all conflict articles associated with an edge as \defn{edge embeddings}.\looseness=-1

\begin{enumerate}[label=\roman*),itemsep=0pt,leftmargin=16pt]
    \item \textbf{Dyadic model} \dyadicmodel{D}: The dyadic model has access to dyadic features only (see top half of \cref{fig:task_construction}D). It takes the node and edge embeddings of a dyad and passes them through multilayer perceptron (MLP) node and edge encoders, respectively. Then, the node and edge embeddings are mean-aggregated at both nodes of the dyad. The averaged embeddings are passed through another MLP and combined through a dot product, which is finally passed through a sigmoid function.\looseness=-1
    \item \textbf{Systemic model} \systemicmodel{S}: The systemic model has access to systemic features only (see bottom half of \cref{fig:task_construction}D). Concretely, it passes all node and edge embeddings through MLP node and edge encoders, except those of the dyad to be classified. Next, the node embeddings are used to initialize a graph isomorphism network \citep[GIN;][]{xu_how_2019} with learnable parameters. In the GIN, edges representing an enemy relationship are weighted by $-1$ and allies by $+1$, with the edge of the dyad being excluded. After a fixed number of message passing steps with the GIN, the resulting node embeddings are mean-aggregated with the edge embeddings, passed through an MLP, and combined as in the dyadic model.
    \item \textbf{Combined} \ceilingmodel{}: The combined model has access to both, dyadic and systemic features. Concretely, it passes all node and edge embeddings, including those of the dyad, through the node and edge encoders and uses all node embeddings to initialize the GIN. The edge of the dyad is of course not weighted to hide the enemy or ally relationship. The rest of the model is identical to the dyadic model.
    \item \textbf{Majority class} (MAJ): This is a majority-class baseline which always predict that two entities are allies.
\end{enumerate}

\paragraph{Hyperparameter settings.}

All models are implemented using \href{https://pytorch.org}{PyTorch}~\citep{pytorch} and the \href{https://docs.dgl.ai}{Deep Graph Library}~\citep[DGL;][]{dgl}. We use the Adam optimizer with $\eta = 0.001, \beta_1 = 0.9, \beta_2 = 0.999$, which have been shown to work well in a variety of settings~\cite{kingma_adam_2015}. We train our models for \num{30} epochs, with early stopping with a patience of \num{3}, and a batch size of $512$. Based on preliminary experiments, we use 2 message passing steps for the GIN. We use ReLU activations for all MLP non-linearities in the network. We ran a grid search to determine the dimensionality of the final layers of node encoder, edge encoder, and edge classifier.\footnote{Details on the grid search and the final hyperparameter values are available on the \href{https://github.com/conflict-ai/conflictwiki}{repository}.}

\paragraph{Data split and training procedure.}
We randomly split the \num{26536} labeled edges of our graph $G$ into a training ($60\%$), validation ($30\%$) and testing set ($10\%$). During training, the entire graph is presented to the model in subgraph batches, but the loss is computed only on the training set edges. This is a form of transductive learning \cite{hamilton_inductive_2017} that eliminates the challenging task of splitting the graph into a separate training and testing graph through sampling (as required by the inductive setting). Moreover, we believe that the transductive setting represents a more realistic scenario, where new entities and conflicts are added to the graph as time progresses and new conflicts erupt.\looseness=-1

\section{Results}
\label{sec:experiments}

The results of our main comparison are shown in the top half of \cref{tab:results_trans_tfidf}. We evaluate results in terms of the \FOne score, which is the weighted average of the precision and recall.
We observe a higher binary \FOne score with the systemic model \systemicmodel{S} (\FOne = \num{0.917}) than with the dyadic model \dyadicmodel{D} (\FOne = \num{0.873}). 
This difference is significant at $p < 0.05$ under a permutation test. We also find that the combined model \ceilingmodel{} achieves \FOne = \num{0.926}, slightly outperforming the models that use only dyadic or systemic features. This asserts that, if our features are to be taken as good representatives of dyadic and systemic correlates, then our results would suggest that conflicts may be better explained by systemic causes rather than dyadic ones.

We conduct two additional analyses to gain further insight into our results. The first is an ablation study of features, to shed light onto the strongest dyadic and systemic correlates (\cref{sec:ablation_features}). The second is a comparative analysis of the article sections that are most similar between allies and enemies (\cref{sec:descriptive_analysis}). 

\subsection{Ablation Study of Features}
\label{sec:ablation_features}

\begin{figure*}[t]
    \centering
    \includegraphics[width=1.0\textwidth]{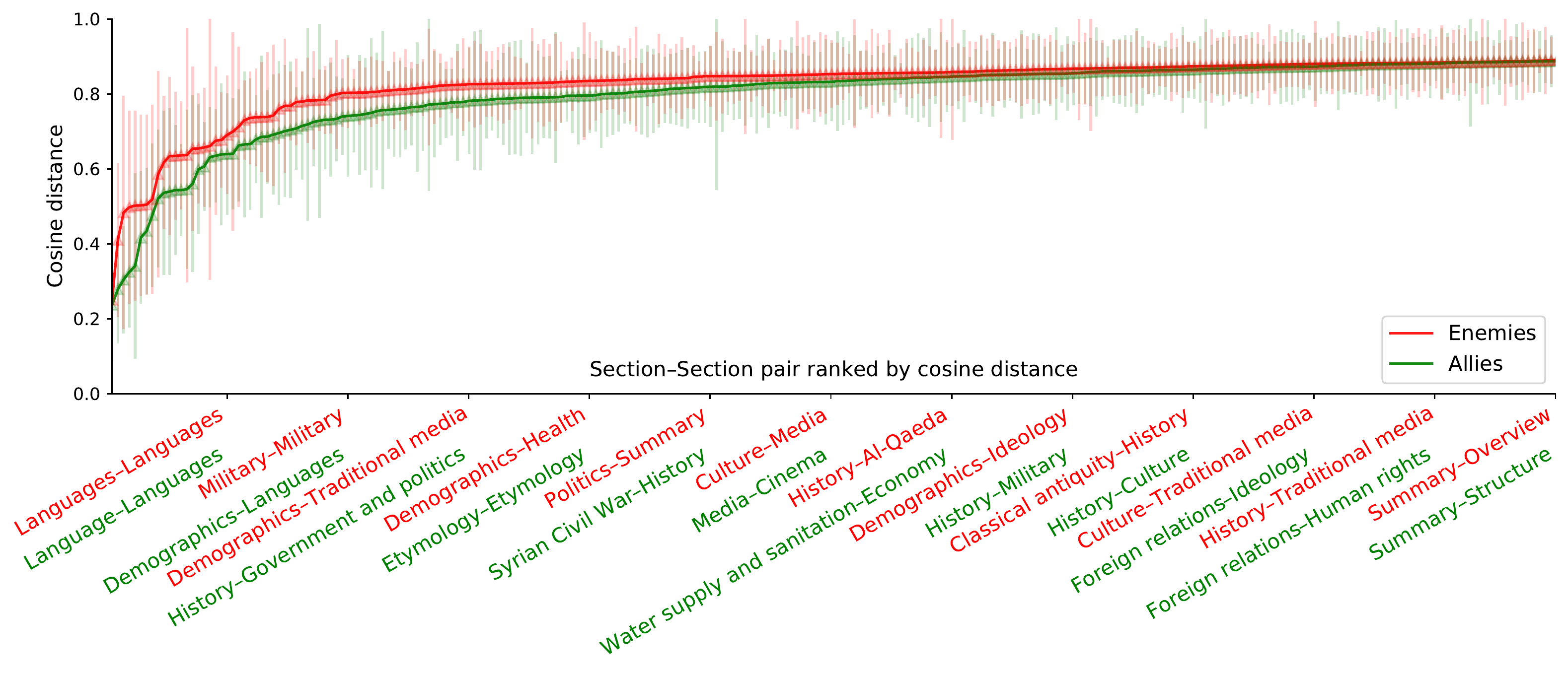} 
    \caption{Top \num{250} most similar pairs of sections of Wikipedia articles between allies (green) and enemies (red), ranked by average cosine distance between tf-idf embeddings (standard deviation in error bars).}
    \label{fig:section_differences}
\end{figure*}

\begin{figure*}[t]
     \centering
     \includegraphics[width=0.95\linewidth]{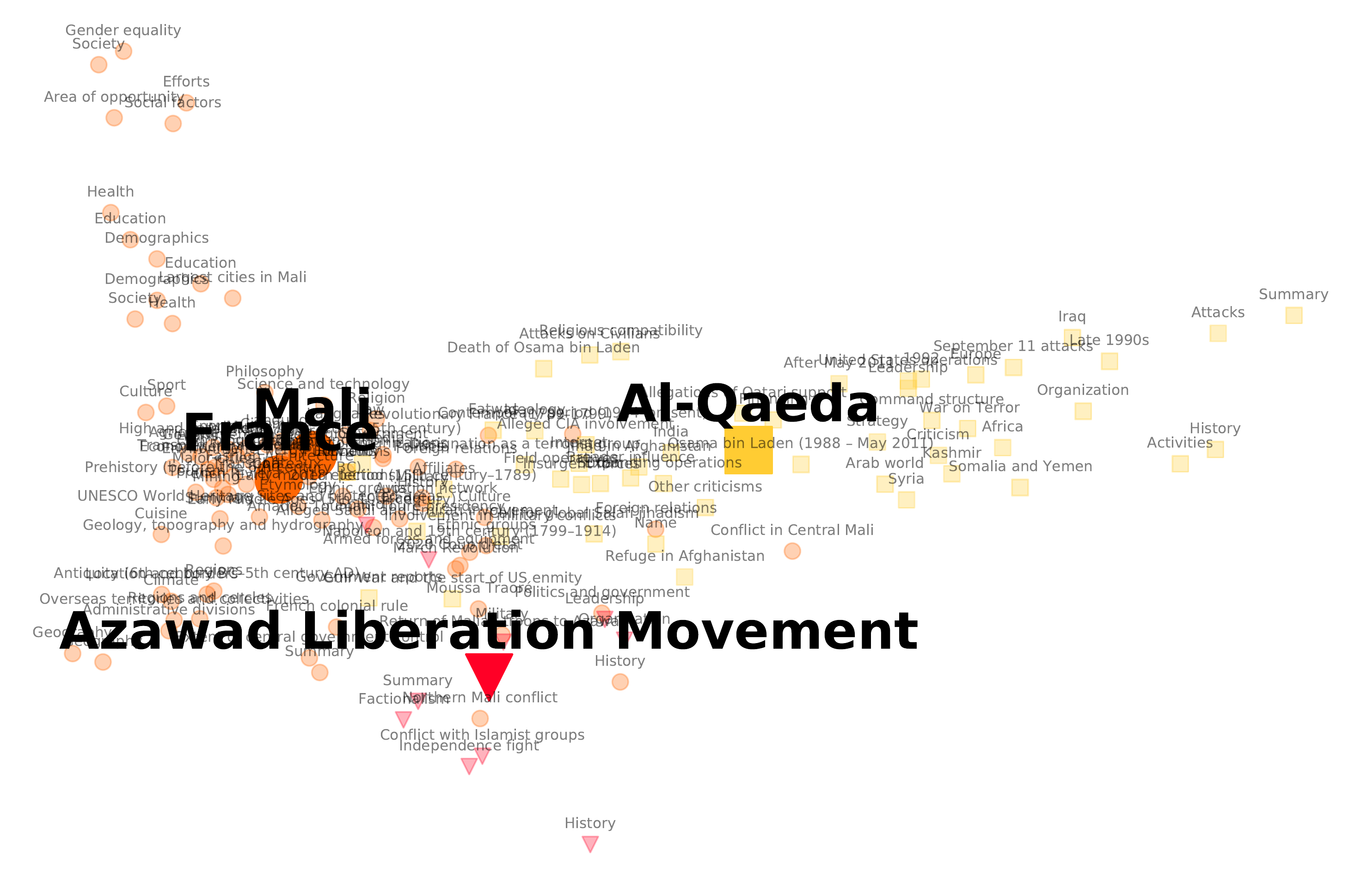} 
     \caption{Top 2 principal components of tf-idf representations of the four entity articles involved in the \href{https://en.wikipedia.org/wiki/Mali_War}{Mali War}; each belligerent is shown with the same symbol. We observe that the allies \Entity{Mali} and \Entity{France} are semantically more similar than enemies.}
     \label{fig:entity_embeddings}
 \end{figure*}

We conduct an ablation study on the individual dyadic and systemic features we defined in \cref{sec:operationalization}. Specifically, we ask the question: out of all dyadic and systemic features, which ones are stronger correlates of militarized conflict? The results are shown in the bottom half of \cref{tab:results_trans_tfidf}.

When leaving out the edge features of the dyad in the dyadic model ablation \dyadicmodel{1}, the \FOne score drops from \num{0.873} to \num{0.836}. This drop suggests that the information contained within the conflict articles is complementary to entity information.
Among the systemic features, we find that the model exploiting only neighboring edge labels \systemicmodel{4} ($\FOne = 0.871$) outperforms both the systemic model that only has access to the node features \systemicmodel{2} ($\FOne = 0.828$), and the systemic model that only has access to the edge features \systemicmodel{3} ($\FOne = 0.779$).
All in all, the results of our systemic feature ablations indicate that, among systemic features, the edge labels appear to be most strongly correlated with conflict; this may give some weight to structural balance theory, which seeks to understand conflict using only the labels of these edges. That said, a stronger correlate is obtained by coupling these labels with other systemic information (as evidenced by the results of \systemicmodel{S}), which seems to indicate that conflict is very much multi-dimensional and cannot be condensed to analyzing binary relationships between entities; in particular, other systemic factors seem to also play a role in conflict.

\subsection{Analysis of Textual Similarity}
\label{sec:descriptive_analysis}

Our results suggest that the dyadic and systemic features we extracted from Wikipedia correlate, to some extent, with whether a pair of entities are allies or enemies. Indeed, some preliminary experiments show that the tf-idf representations of articles and sections are more similar among allies than enemies (for an example, see \cref{fig:entity_embeddings} where the allies \Entity{Mali} and \Entity{France} are closer to each other than to any enemy). To gain further insights into the semantic similarity of allies and enemies, we select the \num{1000} pairs of section titles that most frequently co-occur among allies and enemies and compute the cosine distance of their representations (e.g., Summary--Summary). We plot this in \cref{fig:section_differences}, with ally section pairs shown in green, and enemy ones in red. We find that distance is, on average, lower between allies (mean distance: $0.905$, standard deviation: $0.063$) than between enemies (mean distance: $0.912$, standard deviation: $0.060$) at a significance level of $p < 0.05$ under a $t$-test. This means that entities with similar articles are statistically less likely to appear as enemies in a conflict.\looseness=-1

\section{Related Work}

\paragraph{Entity relationship classification.}

Most work on entity relationship classification is focused on modeling multi-dimensional relations in knowledge bases and ontologies \citep[e.g.,][]{riedel_modeling_2010, miwa_end--end_2016}. The focus of our work is more similar to person-to-person sentiment analysis \cite{west_exploiting_2014} since dyadic relationships are binary. There exist expert-based conflict-cooperation scales such as the Goldstein Scale \cite{goldstein_conflict-cooperation_1992}. Structural balance theory \cite{heider_psychology_1946, cartwright_structural_1956} has been extended to status theory \cite{leskovec_signed_2010} and studied in online discussions by combining signed graphs with sentiment analysis \cite{hassan_detecting_2012, hassan_extracting_2012}. Friend and enemy relations have been studied in novels \cite{iyyer_feuding_2016, srivastava_inferring_2016} and international relations extracted from news \cite{oconnor_learning_2013, tan_friendships_2017, han_no_2019}.

\paragraph{Quantitative conflict studies.}

Consistent with our work, existing empirical studies find evidence for coalescing dyadic and systemic conflict causes \cite{de_mesquita_empirical_1988, midlarsky_systemic_1990, geller_power_1993}. However, empirical studies are limited by availability of text- and graph-based data \cite{harbom_dyadic_2008}. Many machine-extracted (e.g., Europe Media Monitor \citep[EMM;][]{atkinson_creation_2017}) and human-curated (e.g., The Armed Conflict Location \& Event Data Project \citep[ACLED;][]{raleigh_introducing_2010}) conflict event datasets are collections of news articles covering events of daily granularity. Associating events with their overarching long-term conflict and mentioned entities requires complex co-reference resolution \cite{radford_seeing_2020}. The UCDP Global Event Dataset~\citep[GED;][]{sundberg_introducing_2013}, \href{https://ucdp.uu.se/downloads/dyadic/ucdp-dyadic-191.pdf}{UCDP Dyadic Dataset} \cite{harbom_dyadic_2008} and Correlates of War~\citep[CoW;][]{reid_resort_2010} are among the few datasets that associate individual events with overarching conflicts. The UCDP Dyadic Dataset is closest to our dataset, but limited to \num{3000} dyads and does not feature textual descriptions. Related militarized conflict analyses focus on news coverage \cite{west_armed_2017}, interpretable topic models \cite{mueller_reading_2018} and graph neural networks for event detection \cite{nguyen_graph_2018, cui_edge-enhanced_2020}.

\section{Conclusion}

This work explores the extent to which dyadic and systemic features correlate with whether two entities are allies or enemies. Our results suggest that both features are correlated, although, if one is to believe our featurizations and models, systemic features appear to be more correlated. We conduct an ablation study to identify the overall contribution of individual dyadic and systemic features, and a textual similarity study which shows that articles of allies exhibit more similarity than those of enemies.\looseness=-1

\section*{Acknowledgments}

We thank Govinda Clayton, Allard Duursma, Sascha Langenbach and Gokhan Ciflikli for helpful input relating to the background material.

\section*{Impact Statement}

The authors foresee no ethical concerns with the
research presented in this paper.

\bibliography{references,more-refs}
\bibliographystyle{acl_natbib}
\end{document}